\documentclass[journal]{IEEEtran}
\IEEEoverridecommandlockouts
\usepackage{cite}
\usepackage{algorithm,amsmath,amssymb,amsfonts}
\usepackage{algpseudocode}
\usepackage{textcomp}
\usepackage{xcolor}
\usepackage{subcaption}
\usepackage{float}
\usepackage[final]{graphicx}
\graphicspath{ {figures/} }

\DeclareMathOperator*{\maxi}{max}

\def\BibTeX{{\rm B\kern-.05em{\sc i\kern-.025em b}\kern-.08em
    T\kern-.1667em\lower.7ex\hbox{E}\kern-.125emX}}
    
\begin{document}

\title{\huge Resource Management for Blockchain-enabled Federated Learning: A Deep Reinforcement Learning Approach\\}

\author{Nguyen Quang Hieu, Tran The Anh, Nguyen Cong Luong, Dusit Niyato, \IEEEmembership{Fellow, IEEE}, Dong In Kim, \IEEEmembership{Fellow, IEEE}, and Erik Elmroth \\

\thanks{N. Q. Hieu , T. T. Anh, and D. Niyato are with the School of Computer Science and Engineering, Nanyang Technological University, Sinapore  (e-mail: \{quanghieu.nguyen, theanh.tran, dniyato\}@ntu.edu.sg.}
\thanks{N. C. Luong is with the Faculty of Computer Science, PHENIKAA University, Hanoi 12116, Vietnam (e-mail:luong.nguyencong@phenikaa-uni.edu.vn).}
\thanks{D. I. Kim is with School of Information \& Communication Engineering, Sungkyunkwan University, Korea (e-mail: dikim@skku.ac.kr).}
\thanks{E. Elmorth is with Department of Computing Science, Umeå University, Sweden (e-mail: Erik.Elmroth@cs.umu.se)}
}

\maketitle

\begin{abstract}
Blockchain-enabled Federated Learning (BFL) enables mobile devices to collaboratively train neural network models required by a Machine Learning Model Owner (MLMO) while keeping data on the mobile devices. Then, the model updates are stored in the blockchain in a decentralized and reliable manner. However, the issue of BFL is that the mobile devices have energy and CPU constraints that may reduce the system lifetime and training efficiency. The other issue is that the training latency may increase due to the blockchain mining process. To address these issues, the MLMO needs to (i) decide how much data and energy that the mobile devices use for the training and (ii) determine the block generation rate to minimize the system latency, energy consumption, and incentive cost while achieving the target accuracy for the model. Under the uncertainty of the BFL environment, it is challenging for the MLMO to determine the optimal decisions. We propose to use the Deep Reinforcement Learning (DRL) to derive the optimal decisions for the MLMO. 
\end{abstract}

\section{Introduction}

To address the privacy issue of the traditional machine learning, Federated Learning (FL)~\cite{mahan2016} has recently been proposed as an efficient solution that allows mobile devices to cooperatively train a Neural Network (NN) model required by the Machine Learning Model Owner (MLMO), e.g., a server. In particular, the MLMO first transmits the NN model, i.e., global model, to the mobile devices. The mobile devices use their local data to train the model. They then transmit the trained NN models, i.e., the local models, to the MLMO. The MLMO aggregates the local models to the new global model. The MLMO can send the new global model back to the mobile devices for the training. The above steps can be repeated until the global model achieves a certain accuracy. The MLMO pays rewards to the mobile devices for their contributions. 

Since FL communicates the NN model updates, i.e., model's weights, while keeping the data on the mobile devices, it addresses the privacy issue of the traditional machine learning. However, the model updates from the mobile devices and the payment records are still centralized at the MLMO. Therefore, FL faces issues related to the malfunction and the unreliability of the MLMO. In particular, any fault or malfunction of the MLMO, e.g., due to hardware failure or attack, can break the model update communications between the MLMO and the mobile devices. Also, payment records as a financial evidence may be lost or damaged. As a result, the training process needs to be re-initialized~\cite{kim2018}. This calls for a reliable and secure data management scheme for the model updates and the payment records in FL. 



Recently, blockchain has been proposed as an efficient data management approach in FL. Blockchain is used to store and secure reputations of the mobile devices in FL as proposed in~\cite{kang2016incentive}. Moreover, it can be used to prevent malicious mobile devices in FL as presented in~\cite{preuveneers2018chained}. Especically, blockchain is introduced to store the model updates of FL as proposed in~\cite{kim2018}. Indeed, blockchain is considered to be a decentralized database, i.e., a ledger in which transactions are recorded and processed by a number of blockchain nodes, i.e., miners, over the whole network. Such a decentralized database structure addresses the malfunction issue of the MLMO, and it also allows the mobile devices and the MLMO to easily upload and download the model updates. Moreover, blockchain enhances the security and guarantees the data integrity since the transactions must be agreed and verified by the nodes before they are recorded. As such, blockchain can be combined with FL, called \textit{Blockchain-enabled Federated Learning (BFL)}, for efficiently and securely storing the model updates and the payment records.


However, BFL has two major limitations. First, the training latency may increase due to the mining process of the miners in blockchain. The latency increase significantly as the block generation rate decreases. Second, the mobile devices in FL have energy and CPU constraints that may reduce the network lifetime and efficiency of training tasks. To overcome these limitations, the MLMO should decide (i) appropriate amounts of data and energy that the mobile devices use for the training and (ii) the block generation rate in the blockchain
system to minimize the system latency and energy consumption
while achieving the certain model accuracy. However, it is challenging for the MLMO to determine the optimal decisions because the BFL environment, i.e., including the mobile and blockchain environments, are dynamic and uncertain. In this paper, we develop a Deep Reinforcement Learning (DRL)-based scheme to derive optimal decisions for the MLMO without any prior knowledge
of the network.  For this, we first describe the BFL system in which the blockchain system is modeled as an M/M/1 queue system. We then formulate the BFL's problem as a stochastic optimization problem. To solve the problem, we adopt the DRL with Deep Q-Network (DQN)~\cite{mnih2015}. Simulation results show that the proposed DRL scheme outperforms the baseline schemes
in terms of energy consumption, training latency, and cost.

\section{System Model}
\label{sec:system-model}
\begin{figure}[t]
\centering
\includegraphics[scale=0.2]{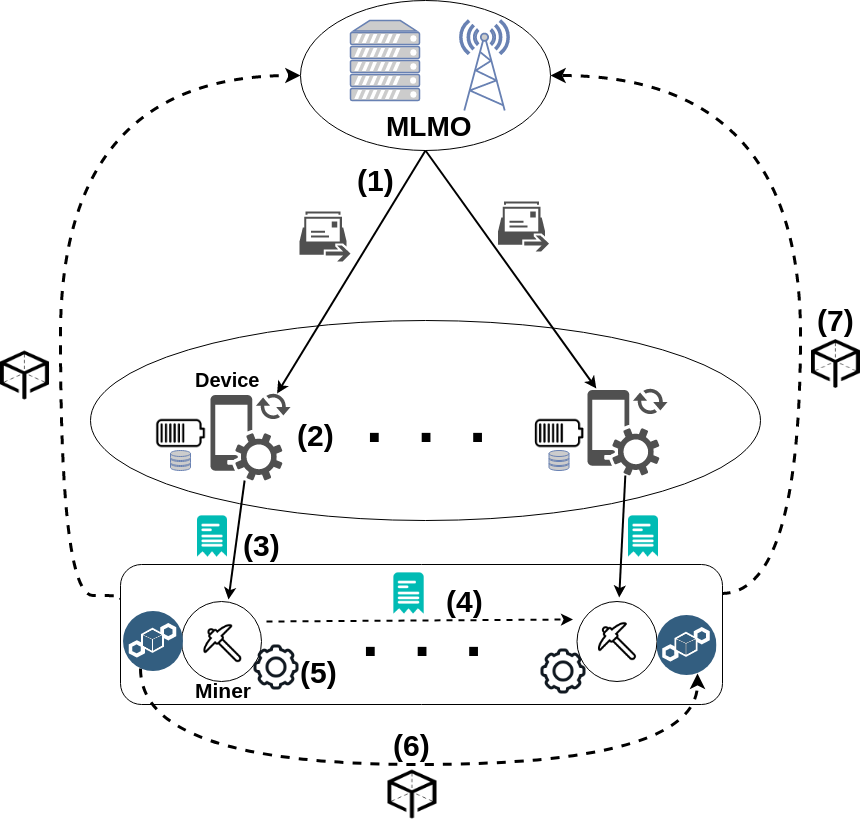}
\caption{\small Blockchain-enabled federated learning network.}
\label{figure:system-model}
\end{figure}

\subsection{System Description}
We consider a BFL network that consists of one MLMO, $N$ mobile devices, i.e., workers, and a blockchain network as shown in Fig.~\ref{figure:system-model} . The MLMO communicates with $N$ devices by using $N$ orthogonal channels through a Base Station (BS). The use of the orthogonal channels is to guarantee that there is no interference among the devices. The MLMO first sends a request to the devices through the BS (Step 1 in Fig.~\ref{figure:system-model}). The request includes an initial global model for training and a record of payments. The payments are monetary rewards as incentives paid to the devices and the miners for their contributions. The payments imply how much resource, i.e., energy and data units, should be used by the devices and what value of blockchain mining/generation rate is used by the miners. Then, each device uses its local dataset, energy and CPU resources to train the global model (Step 2). 
After finishing the training, the device generates a transaction that includes the local model update and the record of payments made by the MLMO in the Step 1. The device sends the transaction to its associated miner (Step 3).
The miner that receives the transactions from their associated devices broadcasts the transactions to the other miners. All the transactions are stored in the miners' queues for being added into blocks. This process is called \textit{cross-verification} (Step 4) which enables the transactions to be synchronized among miners. Next, the miners start mining (Step 5) by using their computation power to solve a cryptographic puzzle as Proof of Work (PoW). The first miner that finds the solution is the mining winner and  is authorized to generate a new block. 
The mining winner then adds a certain amount of transactions into the newly generated block and propagates the block to other miners (Step 6). Other miners receive the block, verify the transactions and add the block to their local blockchain. 
Finally, the MLMO aggregates the transactions from the blockchain and generates a new global model (Step 7). The above steps can be repeated in the next iteration until the global model achieves a certain accuracy. Note that before each iteration, the mobile devices communicate their availability of resources, i.e., energy, with the MLMO. 

\begin{figure}[t]
\centering
\begin{subfigure}[b]{0.8\linewidth}
  \includegraphics[width=0.8\linewidth]{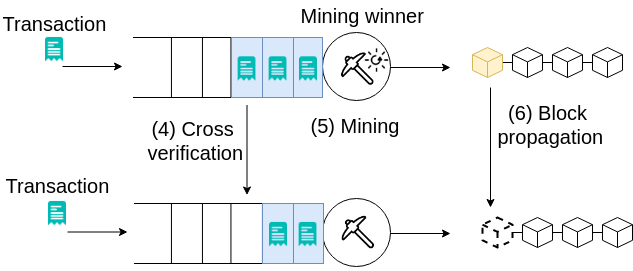}
  \caption{Operation in the blockchain network.}
\end{subfigure}%
\\
\begin{subfigure}[b]{0.8\linewidth}
  \includegraphics[width=0.8\linewidth]{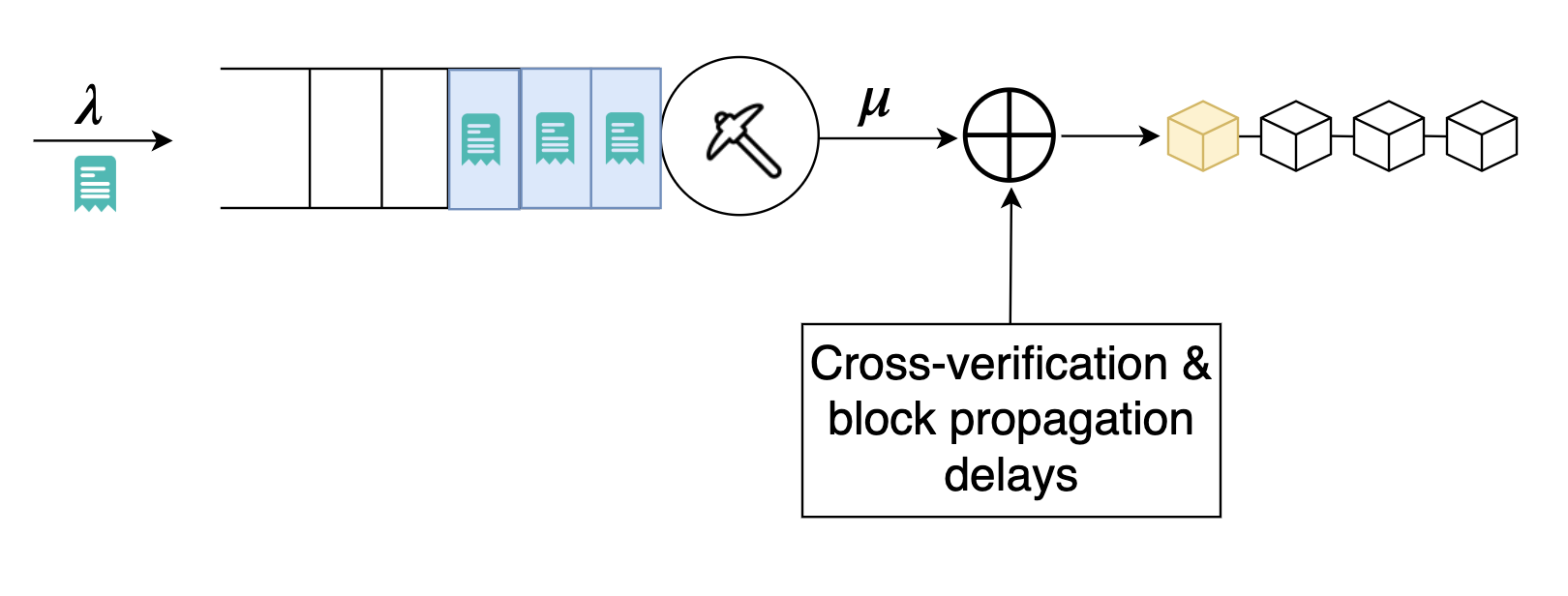}
  \caption{An M/M/1 queue system for blockchain network.}
\end{subfigure}
\caption{\small Blockchain network as an M/M/1 queue system.}
\label{figure:blockchain-model}
\end{figure}

\subsection{Modeling The Blockchain Network}

In fact, the workflow of the blockchain network can be described as in Fig.~\ref{figure:blockchain-model}(a). The input of the blockchain includes transactions transmitted from the devices. The transactions arrive at the network independently and randomly with a constant average rate. Thus, the arrival of the transactions can be considered to be a Poisson process. The output of the blockchain includes new blocks generated by the mining winner. The transactions are stored in the memory of the miners before being added into the next blocks. The blockchain blocks that are generated by the mining is also modeled by a Poisson process. Note that the cross-verification and block propagation are non-stochastic processes~\cite{decker2013}, and they cause extra delays to the system. 

As such, the blockchain network can be modeled as an M/M/$c$ queue~\cite{memon2019} with additional delays as shown in Fig.~\ref{figure:blockchain-model}(b), where M denotes the process that has Markov properties and $c$ is the number of serving nodes. In the queue system, the arrival transactions are stored in a queue before being added into blocks, and the mining winner is the serving node, i.e., $c=1$. 
The mining winner performs a PoW to generate new blocks and adds the transactions into the blocks.
Before being appended to the blockchain, the blocks are suspended by the cross-verification and block propagation delays. 
Because the input and output of the blockchain network are stochastic processes, the number of transactions stored in the queue system at a given time is non-deterministic. The probability that the queue system is in state $m$, i.e., $m$ transactions in the system, can be derived by using queueing theory that is $\mathcal{P}_m = (1- \rho) \rho ^ m$, where $\rho = \lambda / \mu $ is the utilization of the queue system, $\lambda$ and $\mu$ are the arrival rate of the transactions and the block generation rate, respectively. 

\section{Problem Formulation}
\label{sec:problem-formulation}
The problem of the MLMO is to decide (i) the number of data units and the number of energy units from each device $i$ and (ii) the block generation rate of the mining process to achieve the model accuracy target while minimizing the energy consumption and system latency with a reasonable payment. Under the uncertainty of the BFL network, the problem of the MLMO can be modeled as a stochastic optimization problem that is represented by the state space, action space, state transition, and reward function. 
\subsection{State Space}
The state space, denoted by $\mathcal{S}$, is composed of states of the devices and that of the blockchain network. $\mathcal{S}$ is thus defined as $\mathcal{S} = \lbrace
\mathbf{f}, \mathbf{c}, m
\rbrace$, where $\mathbf{f}$ and $\mathbf{c}$ are the vectors including CPU states and energy states of the devices, respectively, and $m$ is the state of the blockchain queue system. In particular, $\mathbf{f} = \left[f_1, \ldots, f_i, \ldots, f_N \right] ^\top$, where  $f_i \in \{ 0, 1, \ldots, F_{max} \}$ is the number of CPU shares that device $i$ uses for training, and $F_{max}$ is the maximum number of available CPU shares. Here, the CPU share refers to the portion of the device's CPU resources that is allocated to the training.
$\mathbf{c}= \left[ c_1, \ldots, c_i, \ldots, c_N \right] ^\top$, where $c_i \in \{0, 1, \ldots, C_{max}\}$ is the energy state of device $i$, i.e., the number of energy units that device $i$ can use for the training, and $C_{max}$ is the energy capacity of the devices. 
\subsection{Action Space}
In each training iteration, the MLMO sends a request that requires $d_i$ data units and $e_i$ energy units from device $i$. The corresponding number of CPU cycles is $f_i^c = \sqrt{\dfrac{\delta e_i}{\tau \nu d_i}}$~\cite{anh2019}, where $\nu$ is the number of CPU cycles required to train one data unit, $\tau$ is the effective switched capacitance that depends on chip architecture of the device, and one energy unit equals $\delta$ Joule (J). Meanwhile, the MLMO needs to determine an appropriate block generation rate of the blockchain. A low block generation rate can decrease the cost paid to the miners, but the system latency is high. In contrast, the high block generation rate decreases the latency, but the cost paid to the miners may be high since the miners consume more computing power to complete the mining process. 

The action space, denoted by $\mathcal{A}$, is thus defined as $\mathcal{A} = \lbrace
\mathbf{d}, \mathbf{e}, \mu
\rbrace$, where $\mu$ is the block generation rate, $\mathbf{d}$ and $\mathbf{e}$ respectively are the vector of data units and that of energy units required by the MLMO for the training. In particular, $\mathbf{d} = [d_1, \ldots, d_i, \ldots, d_N ]^\top$ and $\mathbf{e} = [ e_1, \ldots, e_i, \ldots, e_N]^\top$, where $d_i$ and $e_i$ are the numbers of data units and energy units for device $i$, respectively. Note that the required data and energy resources should not exceed the available resources of the devices, i.e., $d_i \leq D_{max}\ \text{and}\  e_i \leq \min(c_i, E_{max})$, where $D_{max}$ and $E_{max}$ are the maximum numbers of data units and energy units that device $i$ can use for the training, respectively. Also, the number of CPU cycles used for training one data unit does not exceed the available CPU share resource, i.e., $f_i^c \leq \sigma f_i$, where $\sigma$ is the number of CPU cycles corresponding to one CPU share unit.
We have $\mu \in \{\mu_{0}, \mu_0 + 1, \ldots, \mu_{max} \}$, where $\mu_{0}$ and $\mu_{max}$ are the minimal and maximum block generation rates, respectively. We also have $\mu_0 \geq \lambda$ is a constraint that ensures the stable of the blockchain queue system, i.e., the pending transactions in the queue does not increase to infinity.
	
\subsection{State Transition}
The state transition includes the transitions of the states of the devices and the blockchain. The state transition, denoted by $\mathcal{T}(s_t, a_t)$, is defined as $ c_i^{(t+1)} = c_i^{(t)} - e_i^{(t)} + k_i^{(t)}$, $f_i^{(t+1)} \sim U(0, F_{max}) $, and $\mathcal{P}_m = (1 - \rho)\rho^m$,
where $c_i^{(t+1)}$ and $c_i^{(t)}$ are the energy states of device $i$ observed by the MLMO at iteration $t+1$ and $t$, respectively, $e_i^{(t)}$ is the amount of energy that the MLMO requires device $i$ to train the model at iteration $t$, $k_i^{(t)}$ is the amount of energy that device $i$ acquires from wireless charging and $k_i^{(t)}$ is assumed to follow a Poisson distribution~\cite{anh2019}.  Note that the state transition provides the uncertain conditions of the environment and the MLMO has no information about the state transition in advance.

\subsection{Reward Function}

When the MLMO takes an action $a_t \in \mathcal{A}$, it receives an immediate reward. The objective of the MLMO is to achieve a certain model accuracy while minimizing the energy consumption and training latency with a reasonable payment. In particular, the model accuracy is a monotonically increasing function of the total data unit required from the devices for training. Thus, the immediate reward can be defined as follows:  
\begin{equation}
R(s_t,a_t) = \alpha_D \frac{D}{D_{max}} - \alpha_E \frac{E}{E_{max}} - \alpha_L  \frac{L}{L_{max}} - \alpha_I \frac{I}{I_{max}},
\label{eq:reward}
\end{equation} 
where $\alpha_D$, $\alpha_E$, $\alpha_L$, and $\alpha_I$ are the scale factors. The factors are set depending on which subgoals that the MLMO prioritizes. $D = \dfrac{\Sigma_{i=1}^{N}{\eta_i d_i}}{\Sigma_{i=1}^{N}{\eta_i}}$
is the total data unit that the MLMO requires the devices to train their local models and $\eta_i$ is the data quality indicator of device $i$.
$E = \Sigma_{i=1}^{N}{e_i} $ is the total energy unit consumed by the devices for the training.
$L$ is the total latency that is defined as $L = L_{tr} + L_{tx} + L_{blk} $, where $L_{tr}$, $L_{tx}$, and $L_{blk}$ are the latency caused by the training, trasmission, and block generation, respectively. 
$L_{tr}$ is calculated as $L_{tr} = \maxi_{i \in N}(\nu d_i / f_i^c)$~\cite{anh2019}. $L_{tx}$ is the total time for downloading and uploading the models that is determined by $L_{tx} = \dfrac{\delta_d}{W_{dn} log_2(1+\gamma_{dn})} + \dfrac{\delta_d}{W_{up} log_2(1+\gamma_{up})}$, where $W_{up}$ and $W_{dn}$ are the uplink and downlink bandwidths allocated to each mobile device, respectively, $\delta_d$ is the size of the global/local model update, $\gamma_{up}$ and $\gamma_{dn}$ are the SNRs of the uplink and downlink links, respectively. $L_{blk}$ is defined as $L_{blk} = l^{cr}  + l^{bp} + l^{mn}$, where $l^{cr}$,  $l^{bp}$, $l^{mn}$ are the latency caused by the cross-verification, block propagation, and mining, respectively. In particular, $l^{cr}$ and $l^{bp}$ are proportional to the blocks' size that are assumed to be constant\cite{kim2018}, and $l^{mn}$ follows an exponential distribution with the mean of $1/(\mu - \lambda)$~\cite{decker2013}.
$I$ is the total payment including the cost paid to the devices for the training and that paid to the mining winner for the block generation. The cost paid to the winning miner is defined as $\psi_2 \log(1+m)^{-1}$ to normalize the exponent in $\mathcal{P}_m$. It is noted that a high value of $\mu$ reduces $l^{mn}$, but it also reduces $m$ that increases the cost paid to the winning miner. $I$ is thus defined as $I = \psi_1 D + \psi_2  \log(1+m)^{-1}$, where $\psi_1$ and $\psi_2$ are the price factors, $\psi_1 D$ is the cost paid to the devices. Note that components $D$, $E$, $L$, and $I$ of the reward may have different scales, and  to enhance the efficiency of the training, they are normalized by their corresponding maximum values, i.e., $D_{max}$, $E_{max}$, $L_{max}$, and $I_{max}$. 

To maximize the long-term cumulative reward, the MLMO finds the optimal policy $\pi \approx \pi^*$ that allows the MLMO to determine an optimal action $\dot{a_t} \in \mathcal{A}$ given state $s_t \in \mathcal{S}$. For this, the Q-learning algorithm~\cite{watkins1992} can be adopted that constructs and updates a look-up Q-table including Q-values of the state-action pairs, i.e., $Q(s, a)$. 
However, the Q-learning may not be efficient to find an optimal policy when the state and action spaces are large. 
Thus, we propose to use the DQN~\cite{mnih2015}, a combination of Q-learning and deep learning, to find the optimal policy for the MLMO.

\section{Deep Reinforcement Learning Algorithm}

The DQN algorithm uses a Q-network, i.e., a Deep Neural Network with weights $\boldsymbol{\theta}$, to derive an approximate value of $Q^*(s,a)$. 
The input of the Q-network is the states of the MLMO, and the output includes Q-values $Q(s,a; \boldsymbol{\theta})$ of all possible actions. 
The approximate Q-values allow the MLMO to map its state to an optimal action. 
For this, the Q-network needs to be trained to update the weights $\boldsymbol{\theta}$ as follows.  

At the beginning of iteration $t$, given state $s_t \in \mathcal{S}$, the MLMO obtains the Q-values $Q(s, .; \boldsymbol{\theta})$ for all possible actions $a$. 
The MLMO then takes an action $a_t$ according to the $\epsilon$-greedy policy~\cite{watkins1992}.
The MLMO observes the reward $r_t = R(s_t, a_t)$ and next state $s_{t+1}$ and stores the transition $m_t = (s_t, a_t, r_t, s_{t+1})$ to a replay memory $\mathcal{M}$. Then, it randomly samples a mini-batch of the transitions from $\mathcal{M}$ to update $\boldsymbol{\theta}$ as follows:
\begin{equation}
\label{eq: DQN-algorithm}
\boldsymbol{\theta_{t+1}} = \boldsymbol{\theta_t} + \alpha \left[ y_t - Q(s_t, a_t; \boldsymbol{\theta_t}) \right] \nabla Q(s_t, a_t, \boldsymbol{\theta_t}), 
\end{equation}
where $\alpha$ is the learning rate, $\nabla Q(s_t, a_t, \boldsymbol{\theta_t})$ is the gradient of $ Q(s_t, a_t, \boldsymbol{\theta_t})$ with respect to the online network weights $\boldsymbol{\theta}$, and $y_t$ is the target value.
$y_t$ is defined as $y_t = r_t + \gamma \maxi_{a}{Q(s_{t+1}, a; \boldsymbol{\theta_t^-})}$, where $\gamma$ is the discount factor, and $\boldsymbol{\theta_t^-}$ are the target network weights that are copied periodically from the online network weights. The above steps are repeated in iteration $t+1$ to update the weights $\boldsymbol{\theta}$. The training process can be considered to be an episodic task. In each episode, the MLMO averages local model updates from the mobile devices and the episode terminates when the MLMO achieves the certain number of data units $\mathcal{B}$. 


%
%
%
%
%

\section{Performance Evaluation}

In this section, we provide simulation results to evaluate the proposed DQN scheme. 
For comparison, we use the Random policy, the Greedy policy, and the Q-learning\cite{watkins1992} as baseline schemes. 
In the Random scheme, the MLMO selects action $a_t \in \mathcal{A}$ randomly.
In the Greedy scheme, the MLMO always requires the maximum number of data units $D_{max}$ from each device, and then the energy is randomly selected. The simulation parameters are shown in Table~\ref{table:parameters-setting}. In particular, the $\epsilon$-greedy policy with $\epsilon$ decreases from $0.9$  to $0.1$ (from the first episode to the episode 2000) is applied for the action selection to balance the exploration and exploitation. The learning rate is set to $0.001$ to ensure that the training
phase does not miss local minima. Moreover, the first priority of the MLMO is to achieve the high accuracy, and thus $\alpha_D$ is set to the highest value. The subgoals, i.e., low energy consumption and payment, have the lowest priority, and thus $\alpha_E$ and $\alpha_I$ are set to the lowest. 


\begin{table}[t]
\caption{\small Simulation settings}
\label{table:parameters-setting}
\centering
\begin{tabular}{lc}
\hline\hline
{\em Parameters} 			& {\em Value} \\ [0.5ex]
\hline
Maximum CPU share ($F_{max}$) & $3$   \\ 
Energy capacity		           ($C_{max}$)    & $3$   \\ 
Minimum block generation rate     ($\mu_0 $)  & $5$      \\
Arrival rate of the blockchain queue	($\lambda$) & $3$ \\
Scale factors ($\alpha_D, \alpha_E, \alpha_L, \alpha_I$) & $(10, 1, 3, 2)$ \\
 ($\psi_1, \psi_2 $) & $(0.2, 0.8)$ \\
($\sigma, \delta, \tau, \nu$) &  ($0.6$ GHz, $1$, $10^{-28}$, $10^{10}$) \\
$\mathcal{B}$ &  $2000$ data units \\
($W_{up}, W_{dn}$) & $300$ KHz \\
($\gamma_{up}, \gamma_{dn}$) & $10$ dB \\
 $\delta_d$ &$10$ Kb\\
\hline
\end{tabular}
\label{table:parameters}
\end{table}

\begin{figure}[h!]
\vspace{-0.3cm}
\centering
\includegraphics[width=0.36\textwidth]{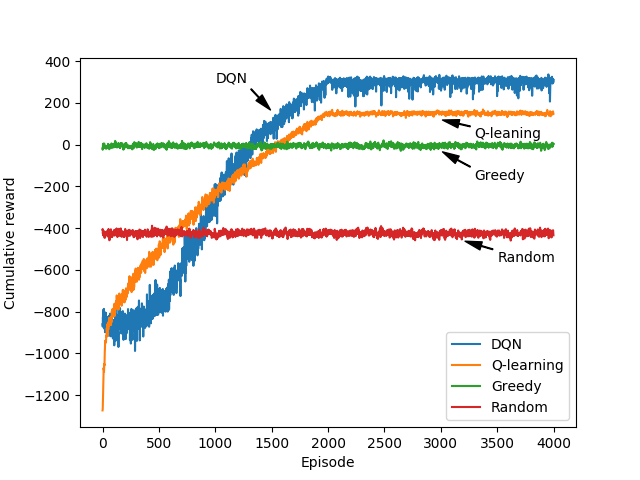}
\caption{\small  Cumulative reward comparison}
\label{figure:reward}
\end{figure}

We first compare the cumulative rewards obtained by the schemes as shown in Fig.~\ref{figure:reward}. As seen, the DQN scheme is able to converge to a cumulative reward value that is much higher than those that of the baseline schemes. In particular, the cumulate reward obtained by the DQN scheme is  $38$\%, $100$\%, and $237$\% higher than those obtained by the Q-learning, Greedy, and Random schemes, respectively. In particular, the reward value obtained by the Greedy scheme is negative. The reason is that the MLMO always takes the maximum number of data units from the devices, and a high amount of energy consumption and system latency may occur that yields the negative reward.
\begin{figure}[h!]
  \vspace{-0.45cm}
\centering
\includegraphics[width=0.35\textwidth]{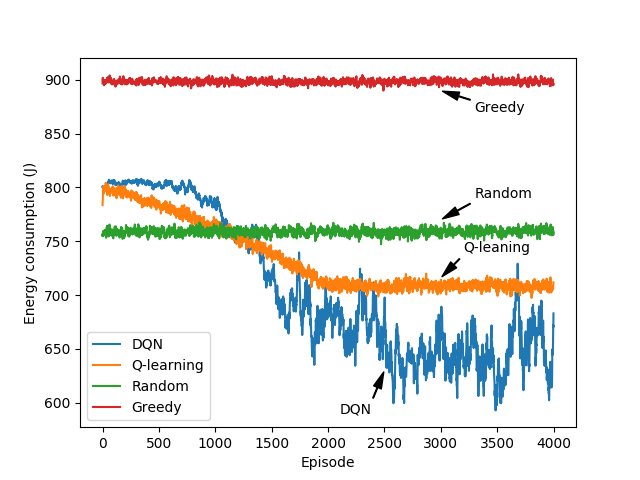}
  \caption{\small Energy consumption comparison}
  \label{figure:energy}
\end{figure}

The high reward improves the energy efficiency of the DQN scheme. As shown in Fig.~\ref{figure:energy}, the DQN scheme significantly reduces the energy consumption compared with the baseline schemes. In particular, the DQN scheme can reduce the energy consumption up to $72\%$ compared with the Greedy scheme. The reason is that the DQN scheme takes the energy cost in the reward function, and optimizing the reward reduces the energy consumption. With the Greedy scheme, the MLMO requires the maximum number of data units, and the devices consume a large amount of energy for the training. 
\begin{figure}[h!]
    \vspace{-0.45cm}
  \centering
  \includegraphics[width=0.35\textwidth]{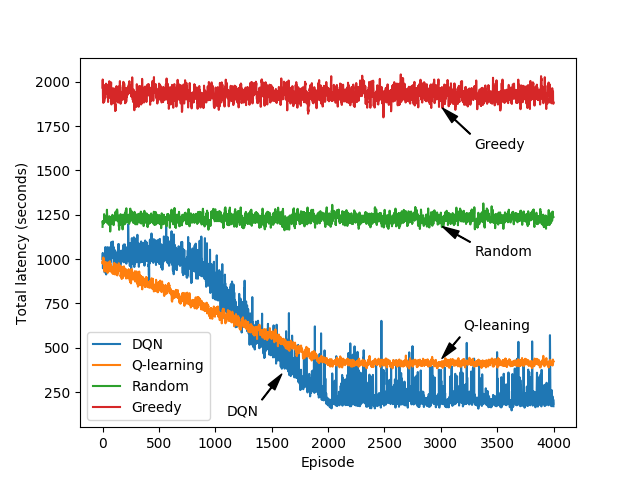}
  \caption{\small Total latency comparison}
    \vspace{-0.33cm}
  \label{figure:latency}
\end{figure}

Fig.~\ref{figure:latency} shows the latency obtained by the schemes. As shown, the DQN scheme outperforms the baseline schemes in term of system latency. In particular, the DQN scheme can reduce the latency up to $12\%$ compared with the Greedy scheme. The reason is that with the DQN scheme, the MLMO decides the amounts of data and energy based on the energy states of the devices. For example, the MLMO takes the small number of data units from the devices with the low energy. This minimizes the training latency. With the Greedy scheme, the latency cost is not considered.  

\begin{figure}[h!]
  \vspace{-0.4cm}
  \centering
  \includegraphics[width=0.37\textwidth]{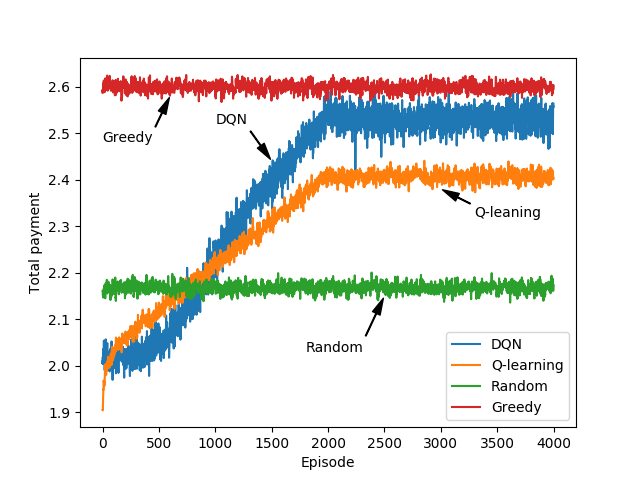}
  \caption{\small Total payment comparison}
  \label{figure:payment}
\end{figure}

Next, we compare the DQN scheme and the baseline schemes in terms of total payment. As shown in Fig.~\ref{figure:payment}, the DQN scheme incurs the total payment lower than the Greedy scheme. However, the DQN scheme still has slightly higher payment than those of the baseline schemes since the DQN scheme needs to balance among the total payment, energy consumption, and system latency. 
\begin{figure}[h!]
\vspace{-0.3cm}
\centering
\includegraphics[width=0.37\textwidth]{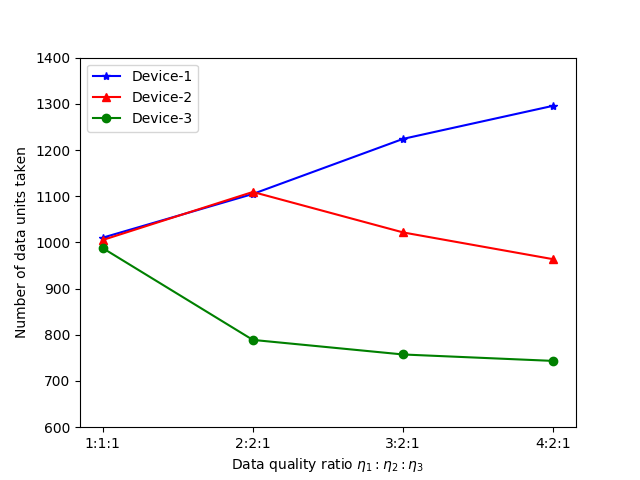}
\caption{\small The amount of data taken as data quality ratio varies}
\vspace{-0.4cm}
\label{figure:data-quality}
\end{figure}

The above experiments are implemented in the scenario in which the devices have the same data quantity. In practice, the devices may have different data quality, and thus it is worth seeing how the MLMO takes the data from each device. Consider the scenario with three devices, we vary the data quality ratio among them by setting $[\eta_1 : \eta_2 : \eta_3]$ to $[1:1:1]$, $[2:2:1]$, $[3:2:1]$ and $[4:2:1]$. As shown in Fig.~\ref{figure:data-quality}, as the data quality ratio is $1:1:1$, the amounts of data taken from the devices are the same. When the data quality of devices 1 and 2 increases, the amount of data taken from these devices is higher than that taken from device 3. Also, as the data quality of device 1 increases to $3$ and $4$, the amount of data taken from this device increases gradually. This simply explains that the MLMO is willing to take more data units from the device with higher data quality such that the MLMO can reach the accuracy target faster.

%
\section{Conclusions}

In this paper, we have presented the DRL scheme for the resource management in the BFL system. We have first described the BFL system in which the blockchain network is modeled as an M/M/1 queue. Then, we have formulated the stochastic optimization problem for the resource management of the MLMO. We have developed a DRL scheme to solve the problem. Simulation results show that the DRL scheme outperforms the baseline schemes in terms of energy consumption and training latency with reasonable cost.


\end{document}